\documentclass{article}

\usepackage[nonatbib,final]{neurips_2022}

\usepackage[utf8]{inputenc} 
\usepackage[T1]{fontenc}    
\usepackage{hyperref}       
\usepackage{url}            
\usepackage{booktabs}       
\usepackage{amsfonts}       
\usepackage{nicefrac}       
\usepackage{microtype}      
\usepackage{xcolor}         

\usepackage{graphicx}
\usepackage{amsmath}
\usepackage[font=small, labelfont=small]{caption}
\usepackage{subcaption}
\usepackage{overpic}
\usepackage{physics}
\usepackage{color}
\usepackage{soul}
\usepackage{wrapfig}
\usepackage[numbers,sort&compress]{natbib}
\usepackage{cleveref}
\usepackage{booktabs}
\usepackage{array}
\usepackage{multirow}
\usepackage{overpic}
\usepackage[final]{pdfpages}

\newcommand{\PreserveBackslash}[1]{\let\temp=\\#1\let\\=\temp}
\newcolumntype{C}[1]{>{\PreserveBackslash\centering}p{#1}}
\newcolumntype{R}[1]{>{\PreserveBackslash\raggedleft}p{#1}}
\newcolumntype{L}[1]{>{\PreserveBackslash\raggedright}p{#1}}

\definecolor{darkgreen}{rgb}{0.0,0.5,0}
\definecolor{darkblue}{rgb}{0.1,0.2,0.8}

\newcommand{\moniker}{GNARF}

\usepackage{paralist}  
\setdefaultleftmargin{2.5em}{2em}{1.7em}{1.5em}{1em}{1em}

\renewcommand{\tr}{{\mathsf{T}}}

\usepackage{xpunctuate}

\providecommand{\wrt}[0]{w.r.t\xperiod}

\crefname{section}{Sec.}{Secs.}
\Crefname{section}{Section}{Sections}
\Crefname{table}{Table}{Tables}
\crefname{table}{Tab.}{Tabs.}
\Crefname{figure}{Figure}{Figures}
\crefname{figure}{Fig.}{Figs.}
\creflabelformat{equation}{#2#1#3}

\newcommand{\mesh}{\mathcal{M}}
\newcommand{\target}{\mathrm{D}}
\newcommand{\canonical}{\mathrm{C}}

\title{Generative Neural Articulated Radiance Fields}

\author{%
  Alexander W. Bergman\thanks{Equal contribution} \\
  Stanford University\\
	\texttt{awb@stanford.edu}\\
  \And
  Petr Kellnhofer$^*$\\
  TU Delft\\
	\texttt{p.kellnhofer@tudelft.nl}\\
	\And
  Wang Yifan$^*$\\
  Stanford University\\
	\texttt{yifan.wang@stanford.edu}\\
	\And
  Eric R. Chan$^*$\\
  Stanford University\\
	\texttt{erchan@stanford.edu}\\
	\And
  David B. Lindell\\
  University of Toronto\\
  Vector Institute\\
	\texttt{lindell@cs.toronto.edu}\\
	\And
  Gordon Wetzstein\\
  Stanford University\\
	\texttt{gordonwz@stanford.edu}\\
}

\begin{document}

\maketitle
\vbox{%
	\vskip -0.15in
	\hsize\textwidth
	\linewidth\hsize
	\centering
	\normalsize
	\tt\href{http://www.computationalimaging.org/publications/gnarf/}{computationalimaging.org/publications/gnarf/}
	\vskip 0.28in
}

\begin{abstract}
	Unsupervised learning of 3D-aware generative adversarial networks (GANs) using only collections of single-view 2D photographs has very recently made much progress.
  These 3D GANs, however, have not been demonstrated for human bodies and the generated radiance fields of existing frameworks are not directly editable, limiting their applicability in downstream tasks.
  We propose a solution to these challenges by developing a 3D GAN framework that learns to generate radiance fields of human bodies or faces in a canonical pose and warp them using an explicit deformation field into a desired body pose or facial expression.
  Using our framework, we demonstrate the first high-quality radiance field generation results for human bodies. Moreover, we show that our deformation-aware training procedure significantly improves the quality of generated bodies or faces when editing their poses or facial expressions compared to a 3D GAN that is not trained with explicit deformations.
\end{abstract}

\section{Introduction}
\label{sec:intro}
Unsupervised learning of 3D-aware generative adversarial networks (GANs) using large-scale datasets of unstructured single-view images is an emerging research area.
Such 3D GANs have recently been demonstrated to enable photorealistic and multi-view consistent generation of radiance fields representing human faces~\cite{Chan2021,deng2021gram,graf,chan2020pi,orel2021styleSDF,Zhou2021CIPS3D,sun2021fenerf}.
These approaches, however, have not been shown to work with human bodies,
partly because learning the body pose distribution is much more challenging given the significantly higher diversity in articulations compared to facial expressions.

Yet, generative 3D models of photorealistic humans have significant utility in a wide range of applications, including visual effects, computer vision, and virtual or augmented reality.
In these scenarios, it is critical that the generated people are editable to support interactive applications, which is not necessarily the case for existing 3D GANs.
While variations of linear blend skinning~\cite{skinningcourse:2014} have been adopted to articulate radiance fields for single-scene scenarios~\cite{peng2021neural,peng2021animatable,su2021anerf,hnerf,2021narf,NeuralHumanPerformer,lin2021efficient,hu2021hvtr,liu2021neural,xu2021sanerf,zhao2021humannerf,weng2022humannerf,jiang2022selfrecon}, it is unclear how to efficiently apply such deformation methods to generative models.

\begin{figure}[t!]
	\includegraphics[width=\textwidth]{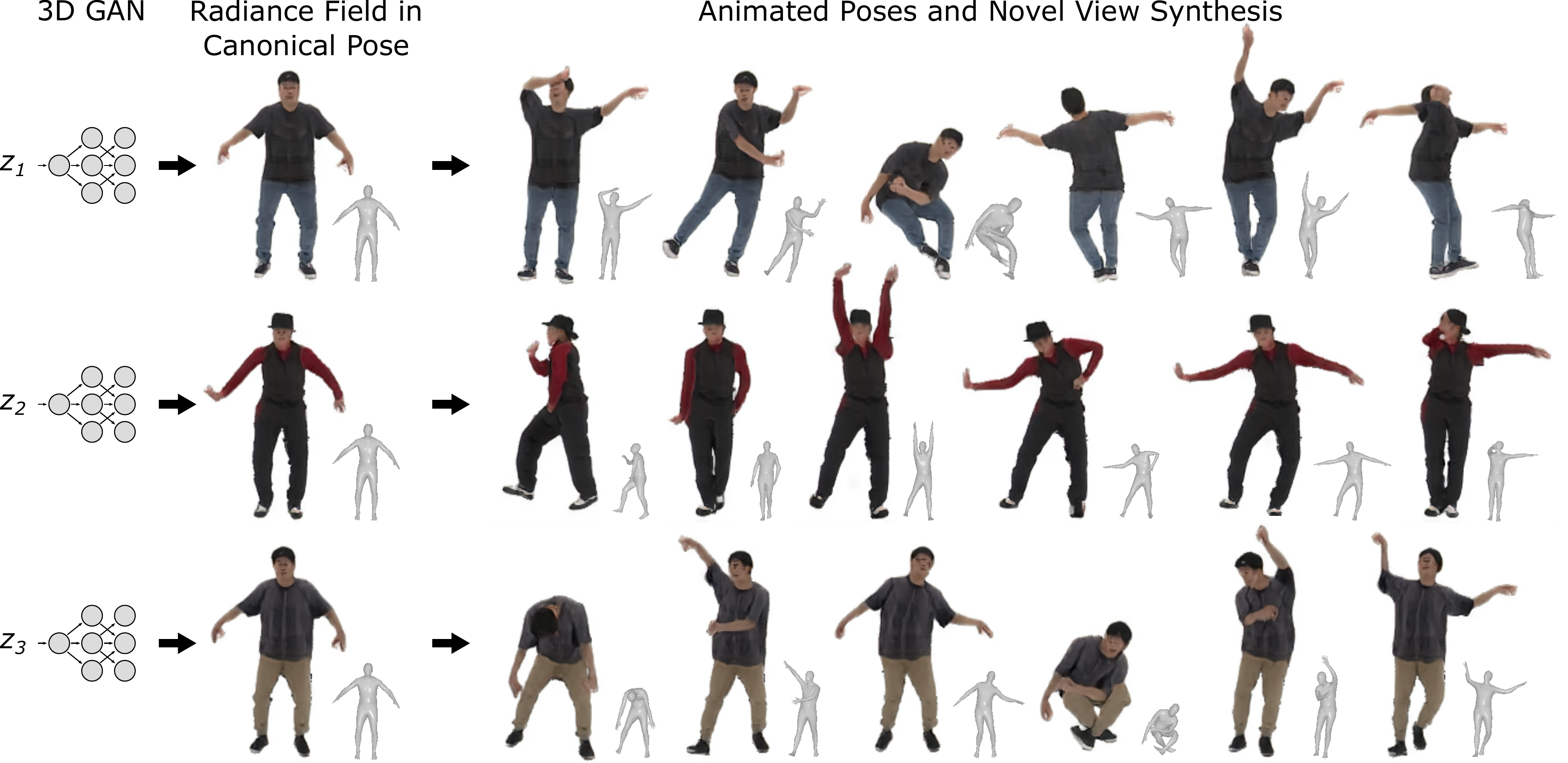}
	\caption{Our method, \moniker, maps a latent space to radiance fields representing human identities. These generated humans can then be animated and rendered from novel views. Qualitative results for our method trained on the AIST++ dataset~\cite{li2021ai} are shown here.}
	\label{fig:teaser}
\end{figure}

With our work, dubbed generative neural articulated radiance fields or \moniker, we propose solutions to both of these challenges.
Firstly, we demonstrate generation of high-quality 3D (i.e., multi-view consistent and geometry aware) human bodies using a GAN that is trained in an unsupervised manner on datasets containing single-view images.
To this end, we adopt the recently proposed tri-plane feature representation~\cite{Chan2021}, which is extremely efficient for training and rendering radiance fields, while being compatible with conventional 2D CNN--based generators, such as StyleGAN~\cite{Karras2020stylegan2}.
While this framework has been successfully demonstrated for faces in prior work, we are the first to adapt it to generating radiance fields of full human bodies.
Secondly, we tackle the editability of the generated radiance fields by introducing an explicit radiance field deformation step as part of our GAN training procedure. This step ensures that the generator synthesizes radiance fields of people in a canonical body pose, which is then explicitly warped according to the body pose distribution of the training data.
We show that this new approach generates high-quality, editable, multi-view-consistent human bodies and that our approach can also be applied to editing faces, increasing the controllability of existing generative models for this task (see Fig.~\ref{fig:teaser}).

To summarize, the contributions of our approach are:
\begin{compactitem}
	\item We present a 3D-aware GAN framework for the generation of editable radiance fields of human bodies. To our knowledge, this is the first approach of its kind.
	\item Our framework introduces an efficient neural representation for articulated objects, including bodies and heads, that combines the recently proposed tri-plane feature volume representation with an explicit feature volume deformation that is guided by a template shape.
	\item We demonstrate high-quality results for unconditional generation and animation of human bodies using the SURREAL and AIST++ datasets and faces using the FFHQ dataset.
\end{compactitem}

\section{Related Work}
\label{sec:related_work}
\paragraph{Articulated 3D Representations.}

Parametric shape templates are one of the most common types of articulated 3D representations adopted in recent neural scene representation and rendering approaches.
These templates, including faces~\cite{Blanz:99,FLAME:SiggraphAsia2017}, bodies~\cite{SMPL:2015}, hands~\cite{MANO:SIGGRAPHASIA:2017}, or a combination of these parts~\cite{SMPL-X:2019}, and even animals~\cite{zuffi20173d},
have been widely utilized for pose estimation and reconstruction, e.g.~\cite{kanazawa2018end,kolotouros2019learning,sanyal2019learning,saito2020pifuhd,kocabas2020vibe}.
Volume deformation is also commonly used in computer graphics, for example using mean value coordinates (MVC)~\cite{ju2005mean} or biharmonic coordinates~\cite{joshi2007harmonic} for shape deformation and editing~\cite{yifan2020neural,sung2020deformsyncnet,jakab2021keypointdeformer,liu2021deepmetahandles,deng2019neural}.
\moniker{} combines parametric template shapes, such as FLAME~\cite{FLAME:SiggraphAsia2017} for heads and SMPL~\cite{SMPL:2015} for bodies, with an intuitive surface-driven volume deformation approach that is both computationally efficient and qualitatively comparable to or better than both skinning and MVC-based deformation.
This provides intuitive editing control for articulated radiance field deformation.

\paragraph{Neural Radiance Fields.}

Coordinate networks, also known as neural fields~\cite{Xie:neuralfields}, have emerged as a powerful tool that enable differentiable representations of 3D scenes~\cite{eslami2018neural,park2019deepsdf,mescheder2019occupancy,michalkiewicz2019implicit,chen2019learning,atzmon2019sal,gropp2020implicit,jiang2020local,davies2020overfit,chabra2020deep,peng2020convolutional,takikawa2021nglod,sitzmann2020siren,martel2021acorn} and learning view-dependent neural radiance fields~\cite{sitzmann2019srns,sitzmann2019deepvoxels,saito2019pifu,liu2019learning,Lombardi:2019,mildenhall2020nerf,martinbrualla2020nerfw,Niemeyer2020CVPR,srinivasan2020nerv,zhang2020nerf,neff2021donerf,liu2020dist,tancik2020fourier,yariv2020multiview,lindell2020autoint,liu2020neural,jiang2020sdfdiff,devries2021unconstrained,kellnhofer2021neural,Reiser2021ICCV,oechsle2021unisurf,yu2021plenoctrees,hedman2021snerg,garbin2021fastnerf,tancik2022blocknerf,mueller2022instant,tewari2020state}.
While initial proposals have focused on static scenarios, recent work has demonstrated successful representations of dynamic scenes~\cite{pumarola2020d,neuralsceneflows,xian2021space,park2021nerfies,park2021hypernerf,tretschk2021nonrigid,Gao-ICCV-DynNeRF,Lombardi:2021}.
Articulated neural radiance fields further extend these approaches by providing editability for neural representations of human heads~\cite{Gafni_2021_CVPR,Wang_2021_CVPR,guo2021adnerf,zheng2022IMavatar,kania2022conerf,zhuang2021mofanerf} and bodies~\cite{peng2021neural,peng2021animatable,su2021anerf,hnerf,2021narf,NeuralHumanPerformer,lin2021efficient,hu2021hvtr,liu2021neural,xu2021sanerf,zhao2021humannerf,weng2022humannerf,jiang2022selfrecon,jiang2022neuman,Knoche20CVPRW} or animals~\cite{yang2022banmo}, often by deforming the underlying radiance fields using traditional 3D morphable models or skeleton-based parameterizations, or alternatively conditioning the radiance field decoder with pose-related parameters.
A more detailed survey of static, dynamic, and articulated neural radiance fields can be found in the recent state-of-the-art report by Tewari et al.~\cite{tewari2021state}. Note that all of these techniques are supervised with scene-specific multi-image data and focus on representing, i.e., ``overfitting'', a single scene. Therefore, it is not easily possible to train these models using unstructured 2D image data and then apply them to generate and edit new and unseen objects or humans. 

\paragraph{Generative 3D-aware Radiance Fields.}

Building on the success of 2D image-based GANs~\cite{radford2016unsupervised,karras2018progressive,karras2019style,Karras2020stylegan2}, recent efforts have focused on training 3D-aware multi-view consistent GANs from collections of single-view 2D images in an unsupervised manner. Achieving this challenging goal requires a combination of 
a neural scene representation and differentiable rendering algorithm. Recent work in this domain builds on representations using meshes~\cite{Szabo:2019,Liao2020CVPR}, dense~\cite{Gadelha:2017,zhu2018visual,henzler2019platonicgan,hologan,nguyen2020blockgan,xu2021volumegan} or sparse~\cite{hao2021GANcraft} voxel grids, multiple planes~\cite{deng2021gram}, fully implicit networks~\cite{graf,chan2020pi,orel2021styleSDF,Zhou2021CIPS3D,sun2021fenerf}, or a combination of low-resolution voxel grids combined with 2D CNN-based image upsampling layers~\cite{Niemeyer2020GIRAFFE,gu2021stylenerf}. Our 3D GAN architecture is most closely related to the recent work by Chan et al.~\cite{Chan2021}, which uses an efficient tri-plane-based volume representation combined with neural volume rendering. We extend this work by including an explicit deformation field in our GAN architecture to model diverse articulations, which allows the generator to synthesize radiance fields of human bodies or heads in a canonical pose while being supervised by 2D image collections that contain arbitrary pose distributions.
Explicitly disentangling radiance field generation and deformation enables us to drastically improve the quality of generated human bodies and faces when their poses or facial expressions are edited.

HeadNeRF~\cite{hong2021headnerf} is related to our approach in that they generate 3D heads conditioned on various attributes. Both HeadNeRF and GNARF condition on identity and facial expression independently, with the former also conditioning on illumination and albedo.
However, to disentangle individual attributes they need to acquire training images of the same person performing various expressions in different lighting conditions.
In contrast, and similar to other 3D GANs, our approach only requires single-view images of different people and can therefore work with readily available 2D image collections.
The recent work by \citet{Grigorev:stylepeople:2021} also generates human bodies. However, their work proposes a 2D GAN that generates textures which are used in combination with a standard articulated template mesh whereas we aim at generating and editing radiance fields using a 3D GAN.
The concurrent work of \citet{Noguchi:narfgan} is closest to ours as the method also includes a radiance field deformation step in a tri-plane-based 3D GAN.
Our evaluation shows superior generation quality and, more importantly, while their approach ties the network architecture to the specific choice of skeleton, our surface-driven volume deformation is agnostic to the particular choice of template and can be used with human bodies, faces, or other object types.

\section{Generative Articulated Neural Radiance Fields}
\label{sec:method}
\moniker{} is a novel general framework to train 3D-aware GANs for deformable objects that have a parametric template mesh, e.g. human bodies and faces.
It builds on the efficient tri-plane feature representation~\cite{Chan2021} for the generated neural radiance field, but additionally applies an explicit deformation which alleviates the requirement for the generator to learn a complicated distribution of articulations.
As a result, the generator automatically learns to generate radiance fields of objects in the canonical pose, which are then warped explicitly to produce target body poses and facial expressions in a fully controllable and interpretable manner.

\subsection{Modeling Articulated Radiance Fields}\label{sec:method-deformation}
We first discuss our approach to modeling and rendering articulated radiance fields, before describing how this is integrated into the 3D GAN in \cref{sec:method-gan}.

\paragraph{Scene representation.}

To represent an object, we leverage the recently proposed tri-plane feature representation~\cite{Chan2021}.
This representation uses three axis-aligned 2D feature planes, each with resolution $N\times N \times C$, where \(N\) and \(C\) denote the spatial resolution and number of channels.
The feature of any 3D point $\mathbf{x} \in \mathbb{R}^3$ is queried by projecting \(\textbf{x}\) onto the planes, retrieving three feature vectors via bilinear interpolation, and aggregating the vectors by summation, i.e., $F\left( \textbf{x} \right) = F_{xy} \left( \mathbf{x} \right) + F_{yz} \left( \mathbf{x} \right) + F_{xz} \left( \mathbf{x} \right)$ where $F_{ij} : \mathbb{R}^3 \mapsto \mathbb{R}^C$ is a function mapping 3D coordinates to features on the $ij$ plane via projection and interpolation.

\paragraph{Articulated deformation.}

We use the deformation function $D : \mathbb{R}^3 \mapsto \mathbb{R}^3$ (detailed later) to warp a coordinate $\mathbf{x}$ from the target (deformed) space into the canonical space. Using a small multilayer perceptron $\textrm{\sc{mlp}} : \mathbb{R}^C \mapsto \mathbb{R}^4$, we convert the deformed 3D feature volume into a neural field of spatially varying RGB colors $\mathbf{c}$ and volumetric densities $\sigma$ as
 \begin{equation}
 \left( \mathbf{c} \left( \mathbf{x} \right), \sigma \left( \mathbf{x} \right) \right) = \textrm{\sc{mlp}} \left( \left( F_{xy} \circ D \right) \left( \mathbf{x} \right) + \left( F_{yz} \circ D \right) \left( \mathbf{x} \right) + \left( F_{xz} \circ D \right) \left( \mathbf{x} \right) \right).
 \end{equation}

There are many possible choices for how to specify the deformation field in an intuitive manner.
For example, linear blend skinning can be used to deform the entire volume ``rigged'' by a skeleton (see e.g.~\cite{skinningcourse:2014}).
While skinning is popular for human body articulations, it cannot explain subtle deformation due to varying facial expressions.
Another option is to use the object-specific template mesh as a cage and apply cage-based deformation for the entire volume using mean value coordinates (MVC)~\cite{ju2005mean}.
However, the high computational cost of evaluating MVCs on the full-resolution grid (see \cref{tab:overfit}) is prohibitive for GAN training and, more critically, this approach generally leads to severe artifacts when the template mesh (accidentally) includes self-intersections (see supplement).

To alleviate these problems, we use an intuitive surface-driven deformation method, which we label as the Surface Field (SF) method.
This method only requires canonical and target template meshes with correspondences, which are readily available for faces~\cite{FLAME:SiggraphAsia2017} and bodies~\cite{SMPL:2015}. These template shapes, in turn, can be driven using skeletons, manual editing, or using keypoints or landmarks that could be detected in and transfered from videos of other people. Therefore, the SF method is generally sufficient to apply to different body parts and it can be intuitively edited in a number of ways, resulting in accurate volume deformation for our class of volumetric models.

The SF approach assigns each 3D coordinate $\mathbf{x}$ to its nearest triangle \(t^{\target}_{\textbf{x}} = \left[ \textbf{v}_{0}, \textbf{v}_{1}, \textbf{v}_{2} \right]\in\mathbb{R}^{3\times3}\) on the target (deformed) mesh. We compute the barycentric coordinates \(\left[ u, v, w \right]\) of the coordinate projected onto this triangle and find the corresponding triangle on the canonical mesh \(t^{\canonical}_{\textbf{x}}\) and its normal \(\textbf{n}^{\canonical}_{t_{\textbf{x}}}\).
The deformed coordinate can then be computed as
\begin{equation}
D\left(\textbf{x}\right) = t^{\canonical}_{\textbf{x}}\cdot\left[ u, v, w \right]^{\tr} + \left< \textbf{x} - t^{\target}_{\textbf{x}}\cdot\left[ u, v, w \right]^{\tr}\hspace{-.25em}, \textbf{n}^{\target}_{t_{\textbf{x}}} \right> \textbf{n}^{\canonical}_{t_{\textbf{x}}},
\end{equation}

The SF approach is very fast to compute and mitigates artifacts from self-intersections of the template shape, thereby combining the benefits of linear blend skinning and MVC-based approaches for the task of radiance field deformation.

\begin{figure}[t!]
	\includegraphics[width=\textwidth]{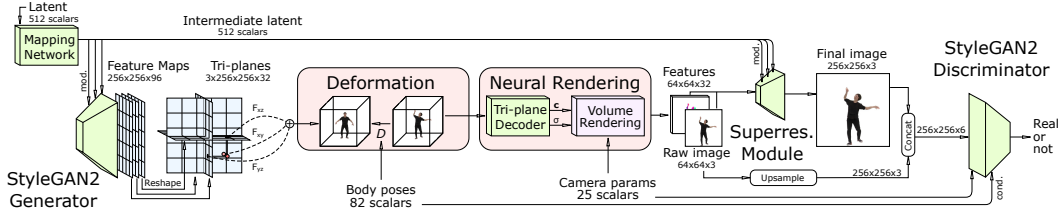}
	\caption{Illustration of the \moniker{} pipeline, including the StyleGAN2 generator, the tri-plane feature representation, feature volume deformation, neural volume rendering, image super-resolution as well as camera view and body-pose conditioned dual discrimination. The resolution of intermediate data and the final image is indicated for experiments with the AIST++ dataset.}
	\label{fig:pipeline}
\end{figure}

\paragraph{Rendering deformed radiance fields.}

We render the radiance field using (neural) volume rendering~\cite{Max:1995,mildenhall2020nerf}. For this purpose, the aggregated feature $F(\mathbf{r})$ of a ray $\mathbf{r}$ is computed by integrating the volumetric features $\mathbf{f}$ and density $\sigma$ as
\begin{equation}
\mathbf{F} (\mathbf{r}) = \int_{t_{n}}^{t_{f}} T \left( t \right) \sigma \left( \mathbf{r} (t) \right) \mathbf{f} \left( \mathbf{r}(t) \right) \textrm{d} t, \quad T \left( t \right) = \textrm{exp} \left( - \int_{t_{n}}^{t_{f}} \sigma \left( \mathbf{r} \left( s \right) \right) \textrm{d} s \right),
\label{eq:volumerendering}
\end{equation}
where $t_n$ and $t_f$ indicate near and far bounds along the ray $\mathbf{r} (t) = \mathbf{o} + t \mathbf{d}$ pointing from its origin $\mathbf{o}$ into direction $\mathbf{d}$. The volume rendering equation (\cref{eq:volumerendering}) is typically approximated using numerical methods, such as the quadrature rule~\cite{Max:1995}.

\subsection{3D GAN Framework}\label{sec:method-gan}

An overview of our 3D pipeline is shown in \cref{fig:pipeline}. Several components, including the StyleGAN generator, the tri-plane representation, the volume rendering, the CNN-based image super-resolution module, and (dual) discrimination are directly adopted from the EG3D framework~\cite{Chan2021}.

Instead of directly generating the radiance field with the target body pose or facial expression, however, \moniker{} is unique in generating the radiance field in a canonical pose and then applying the deformation field discussed in the previous section to warp the feature volume.
We additionally remove the pose conditioning on the generator, and only use camera pose and body pose conditioning in the discriminator. This removes the ability for the generator to incorporate any knowledge about the final view or pose in the canonical radiance field generation, ensuring that the generated results will be robustly animatable beyond just the image rendered at training time. Thus, the generator depends only on the latent code controlling identity, which is input into a StyleGAN2 generator. This architectural choice takes advantage of the state-of-the-art 2D generative model architectures by using them to generate the tri-plane 3D representation. 
The discriminator having access to the camera and body poses ensure that the GAN learns to generate warping accurate to a target pose rather than just being in the correct distribution.
Finally, we adopt a radiance field rendering strategy which samples along each ray inside of an expanded template mesh. This ensures that the integration samples are taken in regions of the radiance field with the most detail and not taken in empty space, simultaneously improving the quality of the generated results and speeding up training.

Additional implementation details, source code, and pre-trained models can be found in the supplement or on our website.

\section{Experiments}
\label{sec:experiments}
We first evaluate the proposed deformation field by overfitting a single representation on a single dynamic full body scene.
Then we apply this deformation method in a GAN training pipeline for both bodies (AIST++~\cite{li2021ai} and SURREAL~\cite{varol17_surreal}) and faces (FFHQ)~\cite{karras2019style}.
Training details and hyper-parameters are discussed in the supplement.

AIST++ is a large dataset consisting of 10.1M images capturing 30 performers in dance motion. Each frame is annotated with a ground truth camera and fitted SMPL body model.
SURREAL contains 6M images of synthetic humans created using SMPL body models in various poses rendered in indoor scenes.
FFHQ is a large dataset of high-resolution images of human faces collected from Flickr.
All images have licenses that allow free use, redistribution, and adaptation for non-commencial use.

\subsection{Single-scene Overfitting}
\label{sec:overfitting}

We compare the proposed surface-driven deformation method, SF, with two alternative methods, MVC and skinning, in a single-scene overfitting task.
MVCs require a set of weights (called the mean value coordinates) to be computed \wrt every vertex of the target mesh \(\mesh^{\target}\) for every sample point. The sample point is then deformed into the canonical pose by linearly combining the vertices of the canonical mesh \(\mesh^{\canonical}\) with these computed weights.
In skinning, the sampling points are deformed to the canonical pose by the rigid transformation of the closest bone as measured by point to line-segment distance.
We find this simplified definition of skinning effective in avoiding blending between two topologically distant body parts (e.g., hand and pelvis) if the starting pose brings them to a geometric proximity.

We select a multi-view video sequence from the AIST++ dataset~\cite{li2021ai} and optimize tri-plane features in the canonical pose using a subset of the views and frames for supervision.
We then evaluate the quality of the estimated radiance field warped into these training views and poses but also into held-out test views and poses.
We apply several modifications to the tri-plane architecture to reduce overfitting; details regarding these changes as well as the selection of training and evaluation views are provided in the supplemental material.

To speed up MVC and SF computation, we decimate the source and deformed SMPL mesh using Quadric Error Metric Decimation~\cite{garland1997surface} in the Open3D library~\cite{Zhou2018} from the original 13,776 faces to 1,376 faces, while tracking the correspondence between the source and deformed meshes.
Nonetheless computing MVC for each deformed pose is still prohibitively expensive for online training (3.7~s per example). We thus precompute the deformation for training and test body poses on a fixed \(16^{3}\) grid and retrieve the deformation for arbitrary sampling points using trilinear interpolation.

\begin{table}[tb]
	\small
	\centering
	\begin{tabular}{lccccccccc}
	  \toprule
		& \multicolumn{3}{c}{Training images} & & \multicolumn{3}{c}{Test images} & & \multirow{2}{*}{Run time [ms] $\downarrow$} \\
		\cmidrule{2-4} \cmidrule{6-8}
		& PSNR $\uparrow$ & SSIM $\uparrow$ & LPIPS $\downarrow$ & & PSNR $\uparrow$ & SSIM $\uparrow$ & LPIPS $\downarrow$ & & \\
		\midrule
		Skinning                  & 18.8 & 0.942 & 0.060 & & 17.9 & \textbf{0.940} & 0.067 & & 95.6 \\
		MVC~\cite{ju2005mean} & 18.1 & 0.937 & 0.067 & & 17.2 & 0.934 & 0.074 & & 0.2 (3782.1)\\
		Surface Field             & \textbf{19.0} & \textbf{0.943} & \textbf{0.058} & & \textbf{18.0} & \textbf{0.940} & \textbf{0.065} & & \textbf{31.6}  \\
		\bottomrule
	\end{tabular}
	\vspace{0.5em}
	\caption{Single-scene overfitting. We evaluate three deformation approaches for the task of estimating a single radiance field in a canonical body pose supervised by a video sequence showing a person from different views and in different poses.
	The SF approach achieves the best quality for both training and unseen test images while being the fastest. Note that the MVC method is only faster than SF when using precomputed grid at the cost of significantly lower deformation accuracy (the runtime without such approximation is reported in parentheses).
The timings are measured to deform a single feature volume on an RTX3090 graphics processing unit.%
	}
	\label{tab:overfit}
\end{table}

As shown in \cref{tab:overfit},
our SF method outperforms the others for both training and test images.
MVC performs worst, partially due to the grid approximation, which is essential in practice.
The skinning method is comparable to SF in terms of image quality but it is 3\(\times\) slower.
Moreover, skinning cannot sufficiently deform subtle facial expressions. Therefore, the SF approach is the most flexible among these deformation methods by being compatible with different human body parts while also offering computational and memory efficiency.

\subsection{Human Body Generation and Animation}
\label{sec:experiment_body}

\begin{figure}[t!]
\centering
	\includegraphics[width=0.85\textwidth]{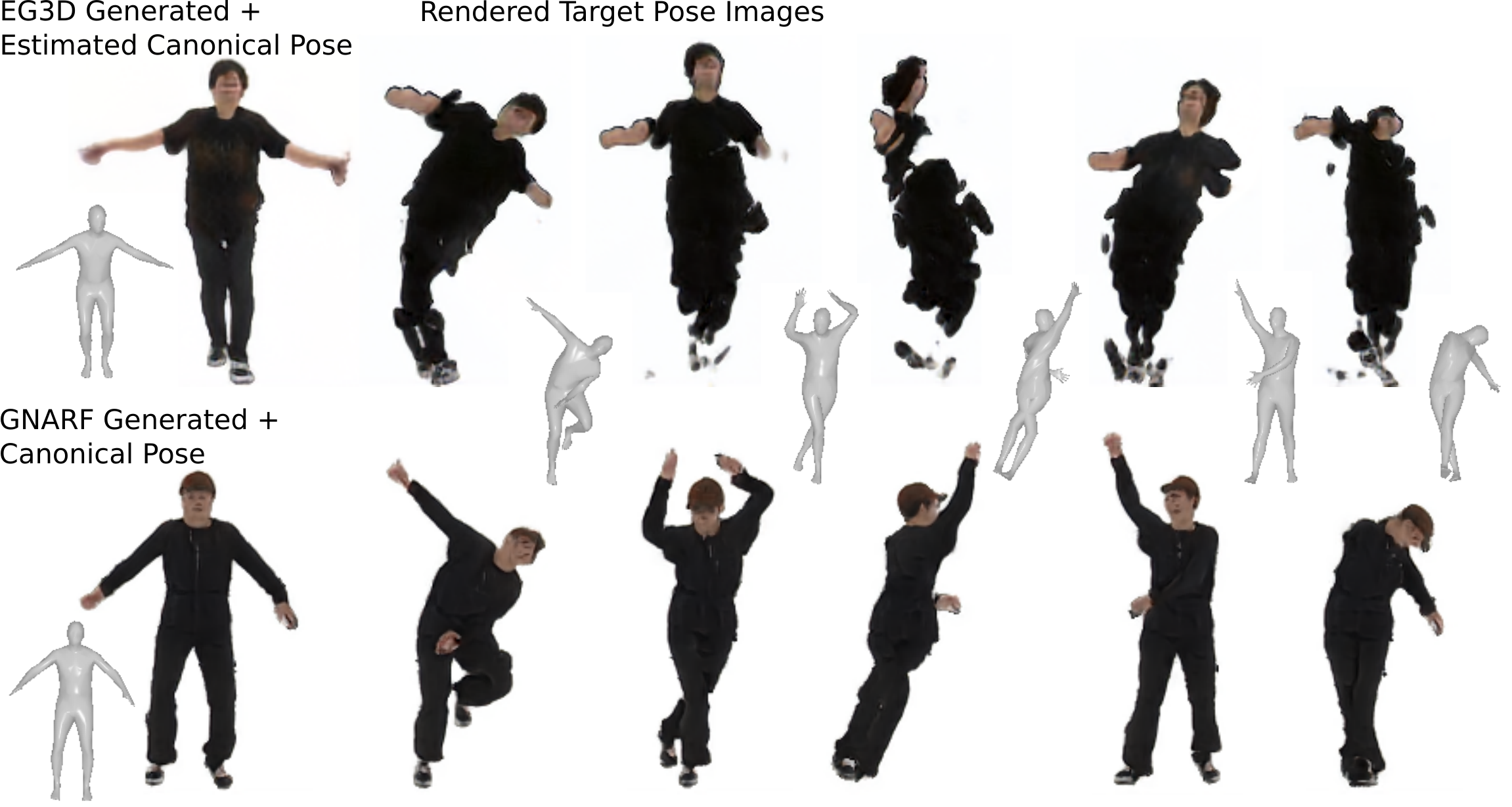}
	\caption{Qualitative comparison of generated target poses using our model vs. warping a pre-trained EG3D model on the AIST++ dataset.}
	\label{fig:qualitative_comparison}
\end{figure}

We now use our SF approach as the deformation method for feature volumes generated by \moniker{}.
Our method is trained and evaluated on the captured AIST++~\cite{li2021ai} and the synthetic SURREAL~\cite{varol17_surreal} datasets.
For both datasets, our method generates high-quality multi-view consistent human bodies in diverse poses that closely match the target pose.

\paragraph{Baselines \& Evaluation.}
Because \moniker{} is the first method to learn a generative model of radiance fields representing bodies, we propose a baseline where we use the original EG3D trained without deformation to generate a feature volume (not in the canonical pose) then warp it into various target poses during inference time using the proposed SF deformation method.
Without the feature volume deformation, the generator is forced to learn to model both identity and pose in its latent space.
As a result, the tri-plane features no longer represent a human body in a consistent canonical pose, but rather match the distribution of poses in the dataset.
The animation of generated bodies is similar to our proposed approach, except that the generated (arbitrarily posed) human body is used as the canonical pose, for which we obtain a SMPL mesh by applying the human shape reconstruction method SPIN~\cite{kolotouros2019learning}.
Additionally, we included the concurrent work ENARF-GAN and its variant ENARF-VAE~\cite{Noguchi:narfgan} in the evaluation whenever a fair and standardized comparison is possible.

We compare FID scores to evaluate the quality and diversity of generated images
as well as the Percentage of Correct Keypoints (PCK) metric to evaluate the quality of the animation.
PCK computes the percentage of 2D keypoints detected on a generated rendered image that are within an error threshold (half the size of the head in the case of PCKh@0.5) of keypoints on a ground truth image in the same pose and view.
We use an off-the-shelf body keypoint estimator~\cite{sun2019deep} trained on MPII~\cite{andriluka20142d} publically available on the MMPose Project~\cite{mmpose2020} to estimate keypoints from a corresponding ground truth and generated image.

\paragraph{AIST++.}
AIST++ is a challenging dataset as the body poses are extremely diverse.
We collect 30 frames per video as our training data after filtering out frames whose camera distance is above a threshold or the human bounding box is partially outside the image.
Then we extract the human body by cropping a $600\times600$ patch centered at the pelvis joint, and resize these frames to $256\times256$.
Since this dataset does not provide ground truth segmentation masks, we use a pre-trained segmentation model~\cite{bazarevsky2020blazepose} to remove backgrounds to stabilize the GAN training.
To speed up training, we use GAN transfer learning~\cite{wang2018transferring}. Rather than initializing our network weights randomly, we begin training from a pre-trained EG3D~\cite{Chan2021} model.
Fine-tuning allows for quicker convergence and saves computational resources during training.
\begin{wrapfigure}{R}{0.24\textwidth}
	\centering
	\includegraphics[width=\linewidth]{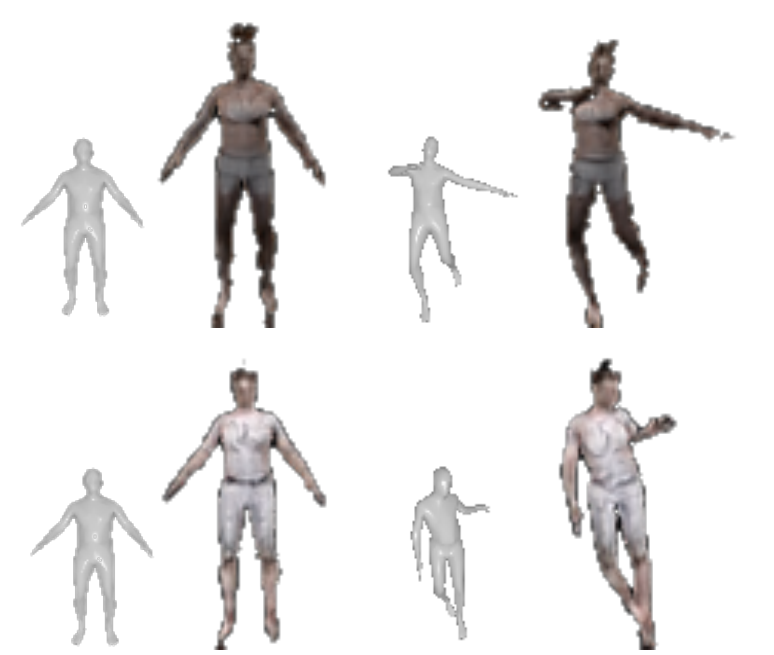}\caption{Example of generated humans in canonical pose and in target pose using model trained on SURREAL dataset.}
\end{wrapfigure}

As shown in \cref{tab:quant},
we can see that our method outperforms the naive reanimation of EG3D by a large margin (see the 4th row vs. the 5th row). Furthermore, our animated method also generates better images than the EG3D baseline that does not support animations.
This is likely due to \moniker{} allowing the generator to focus on generating a specific identity in a canonical pose instead of learning both identity and complex pose distribution in a combined latent space.
Similarly, our model enables high-quality re-animation, as demonstrated by the PCKh@0.5 metric. In \cref{fig:qualitative_comparison}, our method produces significantly better qualitative results than those generated by the baseline.
The baseline results using re-warped EG3D are significantly degraded, since it is difficult to accurately estimate the SMPL mesh from the generated images, which we use as the canonical pose in the deformation function.
Additionally, floating artifacts which exist in the radiance field outside of camera views and make no difference in the conventionally rendered images become visible after being warped can cause false occlusions.
In \cref{fig:teaser}, we show that our method generates bodies in a canonical pose with diverse identities. Additionally, we show that by changing the SMPL parameters, we can drive each radiance field to a desired target pose and render at an arbitrary novel view.

\begin{table}[t]
\small
\centering
\resizebox{\textwidth}{!}{
\begin{tabular}{lccccc}
	\toprule
	& \multicolumn{2}{c}{AIST++ @256$^2$} & \multicolumn{3}{c}{SURREAL @128$^2$} \\ \cmidrule(lr){2-3} \cmidrule(lr){4-6}
	& FID (50k) $\downarrow$ & PCKh@0.5 $\uparrow$ & FID (10k) $\downarrow$ & FID (50k) $\downarrow$ & PCKh@0.5 $\uparrow$ \\
	\midrule
	ENARF-VAE~\cite{Noguchi:narfgan}$^*$ & --- & --- & 63.0 & --- & --- \\
	ENARF-GAN~\cite{Noguchi:narfgan}$^*$ & --- & --- & 21.3 & --- & \underline{0.966} \\
	EG3D (no warping) & \underline{8.3} & --- & \underline{14.3} & \underline{13.3} & ---\\
	EG3D (+ pose est. \& re-warping) & 66.5 & 0.855 & 163.9 & 162.2 & 0.348\\
	\moniker{} & \textbf{7.9} & \textbf{0.980} & \textbf{4.7} & \textbf{5.7} & \textbf{0.999}\\
	\bottomrule
\end{tabular}

}
\vspace{0.2em}
\caption{Quantitative evaluation of our GAN trained on SURREAL~\cite{varol17_surreal} and AIST++~\cite{li2021ai}. Our method only considers foreground images (backgrounds masked). $^*$Metrics provided by authors of~\cite{Noguchi:narfgan} after standardizing the evaluation protocol, which may differ from initial values in their original report.
}
\label{tab:quant}
\end{table}

\paragraph{SURREAL.}
We also test our method on the SURREAL dataset.
The training data is extracted from the first frame of each video in SURREAL's training split.
Each frame is cropped based on the provided segmentation mask from head-to-toe, and resized to $128\times 128$. We filter out the backgrounds and set them to be black.
The SURREAL dataset provides ground truth SMPL parameters and camera intrinsics and extrinsics for each frame.
Similarly to AIST++, we use transfer learning from a pre-trained EG3D model at the appropriate resolution.

In \cref{tab:quant}, we see that our method trained with deformation produces improved FID scores as the version trained without SF feature volume deformation (5th vs 3rd row).
Attempting to deform generated radiance fields results in an immense degradation in quality, shown by both the FID and PCKh@0.5 metrics (4th row).

Additional qualitative results and evaluations are included on the website.

\subsection{Human Face Generation and Editing}
\label{sec:experiment_faces}

Thanks to the surface-driven deformation method, SF, our method can directly utilize expressive parametric face models to drive subtle deformations.
We use the FLAME~\cite{FLAME:SiggraphAsia2017} head model and apply the state-of-the-art facial reconstruction method DECA~\cite{DECA:Siggraph2021} to estimate the flame parameters (100D identity vectors, 6D joint position vectors and 50D expression vectors) from the training dataset.
Since DECA does not account for eyeball movement, we remove the eyeballs from the original FLAME template.
Additionally, we add triangles to close the holes at the neck and mouth area in the FLAME template, which improves the consistency of the SF deformation.
These preprocessing steps produce a mesh with 3,741 vertices and 7,478 faces.
Finally we apply the same decimation method as in the body experiments to obtain a coarse mesh (2,500 triangles), which we use during the GAN training for faster SF deformation.

To train \moniker{}, we start with a conventional EG3D model pre-trained using the FFHQ dataset (see~\cite{Chan2021} for details on the training details and data processing).
We then fine-tune this model using the proposed framework, which includes the SF deformation module. Note that we apply a global scaling and translation to the FLAME template mesh to roughly align it to the faces generated by the pre-trained EG3D model.

\begin{table}
\small
\centering
\resizebox{\textwidth}{!}{
	\begin{tabular}{lccccc}\toprule
		   & FID (500) $\downarrow$ & FID (50k) $\downarrow$ & AED $\downarrow$ & APD $\downarrow$ & ID-Consistency $\uparrow$\\\midrule
EG3D (+3DMM est. \& re-warping)  & \underline{22.9} & \underline{11.6} & 0.29 & \underline{0.028} & \textbf{0.81}\\
PIRenderer~\cite{ren2021pirenderer} & 64.4 & --- & \underline{0.28} & 0.040 & 0.70 \\
3D GAN inversion~\cite{lin:3dganinversion:2022}   & 31.2 & --- & 0.36 & 0.039 & 0.73\\
\moniker{}   			  & \textbf{17.9} & \textbf{6.6} & \textbf{0.23} & \textbf{0.025} & \underline{0.80} \\ \bottomrule
\end{tabular}
}
\caption{Quantitative comparison on FFHQ dataset~\cite{karras2019style}.
}\label{tab:FFHQ_eval}
\end{table}

\begin{figure}[t!]
\scriptsize
\resizebox{\textwidth}{!}{
	\newlength{\imgwidth}
	\setlength{\imgwidth}{0.15\linewidth}
	\setlength{\tabcolsep}{0pt}
	\renewcommand{\arraystretch}{0.8}
	\begin{tabular}{*{5}{C{\imgwidth}}@{\hskip 0.1cm}C{2\imgwidth}}
		\includegraphics[width=\linewidth]{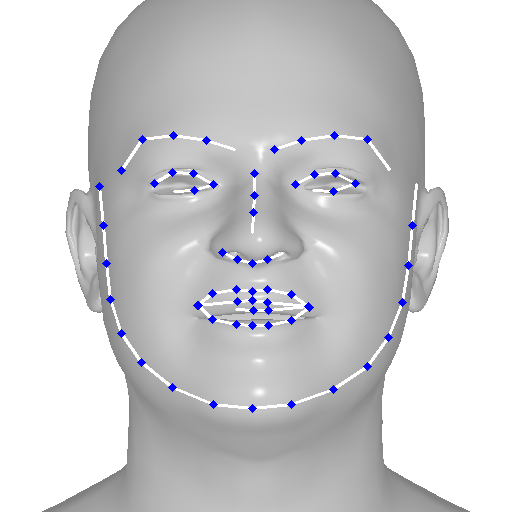} &
		\includegraphics[width=\linewidth]{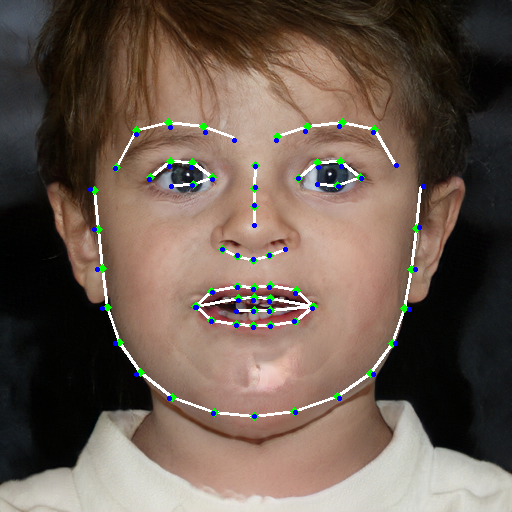} &
		\includegraphics[width=\linewidth]{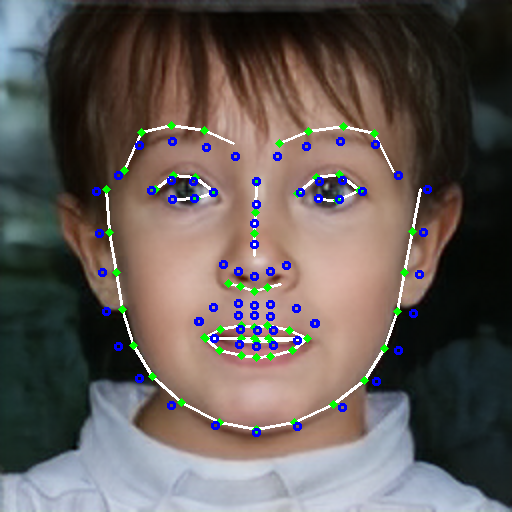} &
		\includegraphics[width=\linewidth]{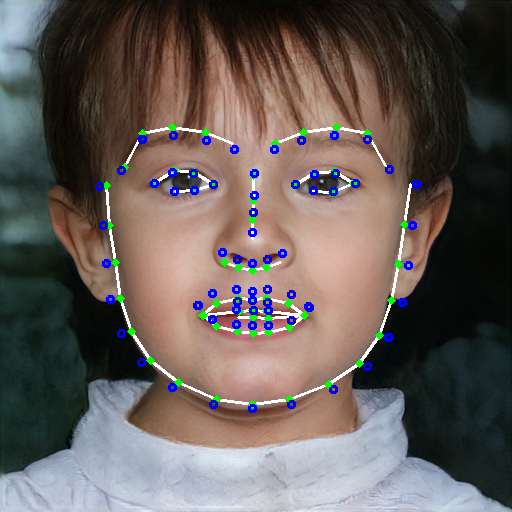} &
		\includegraphics[width=\linewidth]{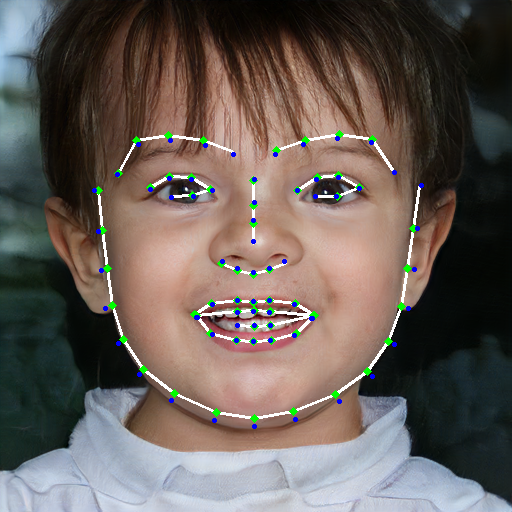} &
		\includegraphics[width=0.5\linewidth, clip, trim=20ex 14ex 20ex 26ex]{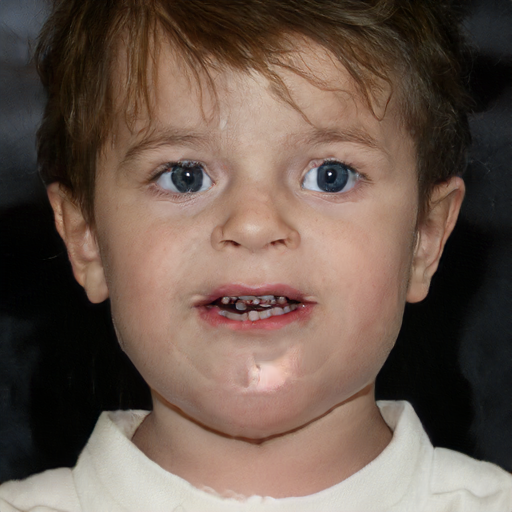}%
		\includegraphics[width=0.5\linewidth, clip, trim=20ex 14ex 20ex 26ex]{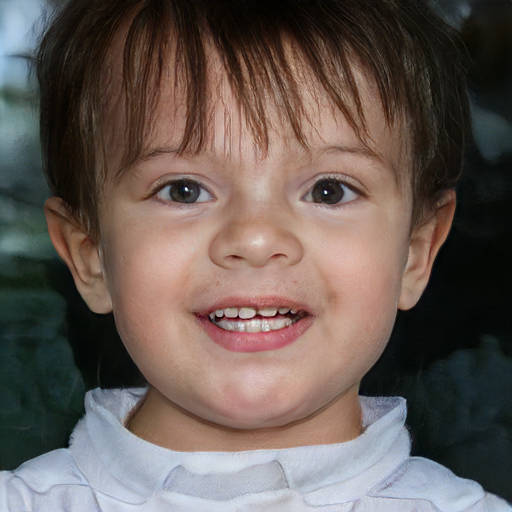} \\
		target face & EG3D+warp & PIRenderer~\cite{ren2021pirenderer} & 3D GAN Inv.~\cite{lin:3dganinversion:2022} & ours &  \multirow{2}{*}{close-ups of EG3D (left) and ours (right)}\\
		& AED=0.36 & AED=0.45 & AED=0.31 & AED=0.25 &
	\end{tabular}
	\let\imgwidth\relax
}
\caption{Deformation fidelity on FFHQ dataset. We plot the target and detected facial landmarks in blue and green, respectively, to visualize the expression fidelity. PIRender~\cite{ren2021pirenderer} and 3D GAN inversion~\cite{lin:3dganinversion:2022} both show large discrepancy to the target face, while the baseline EG3D has strong artifacts from warping (right). Our generated results show high visual quality and they align accurately with the driving deformation.}
\label{fig:expression}
\end{figure}

For evaluation, our first baseline is the original EG3D model with re-warping at inference time, as described in \cref{sec:experiment_body}.
We also include two state-of-the-art facial reenactment methods, PIRenderer~\cite{ren2021pirenderer} and 3D GAN inversion~\cite{lin:3dganinversion:2022}.
Several metrics are used by our quantitative evaluation.
FID (500) follows the evaluation protocol of \citet{lin:3dganinversion:2022}, where the ground truth dataset consists of 500 randomly sampled identities and the test dataset is constructed by animating the ground truth using randomly sampled target poses.
FID (50k) follows the protocol in EG3D, where the entire FFHQ dataset is treated as the ground truth and the test dataset includes 50k generated images using randomly sampled latent vectors, camera poses and FLAME facial parameters.
Following \cite{lin:3dganinversion:2022}, we evaluate the fidelity of the animation with the Average Expression Distance (AED) and the Average Pose Distance (APD) computed using DECA estimation, as well as the identity consistency based on a face recognition model~\cite{deng2019arcface}.

\begin{figure}[t]
\centering
\includegraphics[width=\textwidth]{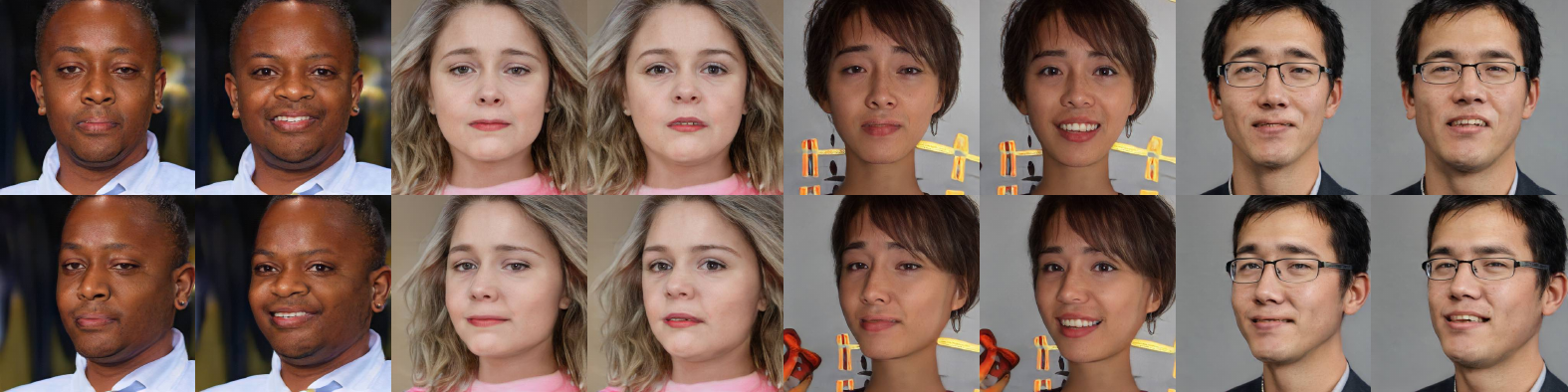}
\caption{Qualitative results on FFHQ dataset. We show each identity in their canonical (left) and a different (right) expression from two different camera poses. %
Our method shows excellent multi-view consistency.}\label{fig:faces}
\end{figure}

As shown in \cref{tab:FFHQ_eval}, our method is superior to the baseline methods in both FIDs and in the edited fidelity, and comparable to the baseline EG3D in terms of identity consistency.
The advantage of our method in editing ability is clearly demonstrated in \cref{fig:expression}: our method reproduces the target expression more accurately when compared to the face reenactment methods and mitigates the warping artifacts present in the baseline EG3D.
In \cref{fig:faces}, we further demonstrate the quality of our results.
Notice that even though the FLAME model does not include teeth, the SF deformation guides the neural radiance field to construct teeth consistently at the correct location.

\section{Discussion}
\label{sec:discussion}
\paragraph{Limitations and future work.}

Our work is not without limitations. The level of detail in the generated bodies, for example, is relatively low. This is partly due to the limited resolution of the training data in the SURREAL and AIST++ datasets, but also due to the limited resolution that the tri-plane representation offers for any one body part, such as the face. An interesting avenue of future work could include the exploration of adaptive radiance field resolution for human bodies, allocating more resolution to salient parts (see e.g.~\cite{fruehstueck:2022}). The image quality of the generated bodies is currently on par with simpler approaches that only need to generate an RGB texture on the SMPL mesh~\cite{Grigorev:stylepeople:2021}.
Yet, details in faces and hair cannot be handled by a texture-generating approach and we expect the quality of generative radiance fields to surpass that of simpler alternatives with increasing and perhaps adaptive radiance field resolutions.
Another bottleneck of our current framework is the challenge of being able to work with large-scale datasets showing a diversity of visible humans.
In-the-wild datasets, such as MSCOCO~\cite{Lin:MSCOCO}, do contain this diversity but also contain a significant amount of occlusion and detailed backgrounds, requiring the generator to spend capacity on modeling the distribution of backgrounds and occlusions themselves. An interesting avenue of future work consists of explicitly modeling occlusion and background in the GAN training pipeline independently from identity and pose. Currently, the background is not modeled in the generative framework and requires pre-processing of the dataset to separate foreground from background.
Additionally, pose estimation from in-the-wild images is still not very accurate, degrading the quality of our deformation function.
On the other hand, large-scale custom curated datasets such as the one used in InsetGAN~\cite{fruehstueck:2022} are not publicly available.
Finally, the deformation we use as part of the 3D GAN training is limiting in several ways: it does not allow for topology changes, it does not prevent solid parts like teeth or eyeglasses from being stretched, and the deformation quality degrades for points far from the surface. In general, using the surface of a parametric mesh to guide an volume may not be the optimal choice for more complex volumetric scenes. More advanced deformation methods, perhaps including kinematic constraints, could be an interesting direction of future research.

\paragraph{Ethical considerations.}
GANs could be misused for generating edited imagery of real people. Such misuse of image synthesis techniques poses a societal threat, and we do not condone using our work with the intent of spreading disinformation. We also recognize a potential lack of diversity in our results, stemming from implicit biases of the datasets we process.

\paragraph{Conclusion.}
Our work takes important steps towards photorealistic 3D-aware image synthesis of articulated human bodies and faces with applications in visual effects, virtual or augmented reality, and teleconferencing among others.

\begin{ack}
	Alexander W. Bergman was supported by a Stanford Graduate Fellowship.
	Gordon Wetzstein was supported by NSF Award 1839974, Samsung, Stanford HAI, and a PECASE from the ARO.
	We thank Connor Lin for helping standardize comparisons to ~\cite{lin:3dganinversion:2022}. We thank Atsuhiro Noguchi and the authors of ~\cite{Noguchi:narfgan} for helping in standardization of the comparisons of our methods.
\end{ack}

{\small
	\bibliographystyle{unsrtnat}
	\bibliography{references}
}



# 1 Implementation Details

Applications of our method to human bodies and faces share the same framework, but differ in a few implementation details. For clarity, in the following section we describe the implementation details for GNARF applied to human bodies and outline the differences for faces in Sec. 2.3.

## 1.1 Generator and tri-plane representation

We use the generator architecture from EG3D [1], which is built on top of the public StyleGAN2 [2] architecture located at https://github.com/NVlabs/stylegan3 (this StyleGAN3 repository contains backward compatibility for StyleGAN2). The generator is composed of four components: a mapping network, a convolutional backbone, an MLP decoder, and a convolutional super-resolution module.

The generator is conditioned on a 512-dimensional Gaussian noise input using a two-layer mapping network of 512 hidden units. We do not condition the generator on either camera pose or body pose. The mapping network produces a 512-dimensional latent code. This latent code modulates the layers of a StyleGAN2-based convolutional backbone, which produces a 96-channel $256 \times 256$ feature image. This is reshaped into three axis-aligned tri-planes, each of shape $256 \times 256 \times 32$. This architecture is trained from scratch rather than using a pre-trained StyleGAN2 network.

The MLP decoder which operates on top of sampled plane features consists of a single hidden layer of 64 units. The decoder maps the 32-dimensional sampled plane feature to a 33-channel feature consisting of a scalar density and 32-dimensional feature. These are integrated per the volume rendering equation (Eq. 3 in the main paper) to obtain a $64 \times 64 \times 32$ feature image, where $64$ is the spatial resolution and $32$ is the number of channels.

As in EG3D, a separate super-resolution module (implemented as CNN) up-samples and converts the feature images to the final RGB output. The final resolution of the output is $128^2$ for SURREAL and $256^2$ for AIST++ respectively. As in EG3D, this module is implemented with two StyleGAN2 convolutional blocks, with channel depth of 128 and 256 respectively.

## 1.2 Deformation and volume rendering

The SF deformation is performed using a simplified version of the SMPL mesh. As described in the main text, this simplified version is obtained using Quadratic Error Metric Decimation [3] in the Open3D library [4] to reduce SMPL from 6890 vertices and 13,776 faces to 690 vertices and 1376 faces.

We perform volume rendering in the canonical space with 64 uniformly-spaced samples plus 64 additional samples based on importance sampling [5] per ray. Additionally, we sample rays only inside an expanded version of the simplified SMPL mesh. The mesh is expanded using a growth offset parameter $g$, which controls the new position of a vertex $v$ with vertex normal $n$, by moving $v$ to $\hat{v} = v + gn$. We use $g = 0.05$ during training.

As described in the main paper, deformation methods such as Mean Value Coordinates (MVC) [6] and skinning introduce large artifacts when the template mesh (accidentally) intersects itself or comes very close to doing so. This happens often in practice, even with the unmodified SMPL mesh, due to imperfect SMPL parameter estimation or a subject actually touching somewhere on their body.

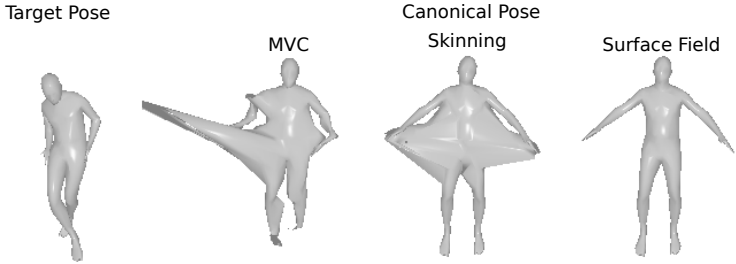

Figure 1: Deforming the vertices of the target pose on the left into the canonical pose results in artifacts for the MVC and skinning deformation method, while the surface field deformation method results in a perfect deformation by construction.

This is shown in Fig. 1: the hands coming close to the body result in large deformation artifacts in the canonical space for MVC and skinning. In the case of MVC, this is mainly attributed to MVC being non-local - a point has non-zero weight w.r.t. *all* the vertices in the driving mesh. This is worsened by the grid approximation mentioned in the main paper (this is necessary to enable feasible training time using MVC): the grid approximation to MVC does not guarantee that surface points remain on-surface after deformation (unlike the full MVC computation) since the rapid change in MVC weights near the surface cannot be sufficiently captured in feasible grid resolution. Similarly, in the skinning method, when parts of the driving mesh get close together, individual vertices may become closer to a different bone than the one which they are a part of. This causes them to deform to incorrect places in the canonical space. This artifacts are significantly mitigated when using the proposed surface field (SF) deformation, as SF is very local (in contrast to MVC) and the lookup of the closest triangle is less error-prone than of the closest bone.

### 1.3 Discriminator

**Dual discrimination.** Similarly to EG3D, we use dual discrimination to ensure consistency between the raw neural rendering and the final super-resolved output. As in EG3D, we concatenate a resized copy of the raw neural rendering to the super-resolved input to form a 6-channel discriminator input tensor. This raw neural rendering consists of the first three channels of the rendered feature image.

**Discriminator pose conditioning.** As in EG3D, we condition the discriminator on the camera pose via a mapping network that modulates the layers of the discriminator. Unlike EG3D, we additionally condition the discriminator on the expected body pose / facial expression by concatenating the body/facial pose parameters (SMPL or FLAME parameters) to the camera parameters as input to the mapping network.

By conditioning on body / facial poses, we give the discriminator the ability to ensure that the applied deformation matches the specified pose. Empirically, we found that corrupting the poses with 1 standard deviation of Gaussian noise before passing them as input to the discriminator aided training convergence. We hypothesize that in the absence of noise, the discriminator was able to overfit on the specific poses and cameras of the ground truth dataset, which destabilized training.

Note that unlike EG3D, which conditions the generator on the camera parameters for training with FFHQ *(modeling pose-correlated attributes)* we do not condition the generator with camera pose. We also do not condition the generator with body pose or facial expression. This is done in order to ensure that the generator learns to generate a body/face in the canonical space which is robust to custom deformations, rather than one which is specific for a warping or camera viewpoint.

### 1.4 Training

Many of our training hyperparameters are adopted from those of EG3D and StyleGAN2: generator learning rate (0.0025), discriminator learning rate (0.002), batch size (32), blurring images (GT and generated) over the first 200K iterations, and R1 regularization [7]. As recommended, the gamma parameter of R1 regularization is tuned according to the dataset: FFHQ: $\gamma = 1$; SURREAL: $\gamma = 1$; AIST $\gamma = 4$.

We use 8 Tesla V100 GPUs, training each model for roughly 2 days.

## 2 Additional Results and Evaluation Details

### 2.1 Single-scene overfitting.

**Data pre-processing.** The single-scene overfitting experiment is conducted using the sequence *gBR_sBM_cAll_d04_mBR0_ch01* from the AIST++ dataset [8]. We extract all 719 frames sampled at 60 Hz from each of the 8 available cameras and we crop the human body using the same procedure as in the GAN training. We skip the camera number 4 because the annotation in the dataset does not match the video. We hold out the cameras number 2 and 7 for testing and we train our models using 30 frames uniformly sampled from the remaining six cameras.

**Model.** We use the same triplane representation as in our GAN experiments with a few modifications to avoid overfitting to the sparse training data. First, we limit the capacity of the decoder network by reducing the latent space size from 64 to 32 and by reducing the resolution of the triplanes from 256 to 128. Next, we increase the number of samples per ray to 128 and disable the second stage of importance sampling. Finally, unlike in the GAN setup, we do not remove background from the training images. Instead, we train a representation of the entire scene as is common in other overfitting papers [9]. To avoid mixing of the static background and dynamic foreground in the neural representation and to allow for efficient sampling of both regions with different depth ranges, we include a separate identical triplane representation for the background. We use the accumulated optical density from the foreground network to alpha-blend the foreground image over the background image. No ground-truth foreground masks are used during the training.

We utilize the same model for deformation function $D$ implemented using mesh skinning, Surface Field and MVC. For skinning and Surface Field, we compute the transformations on-the-fly. However, this is not feasible for the relatively slow MVC computation. Therefore, we precompute the MVC transformations for $16{\times}16{\times}16$ points uniformly sampled withing a bounding cube of the human body for all training poses, and we sample them during training using a trilinear interpolation.

We train our model for 500 000 steps with Adam optimizer [10] and step size of 0.002 and we use L2 loss to supervise the training at $128{\times}128$ resolution with batch size of 3 images on Nvidia RTX3090 graphical processing unit.

**Evaluation.** We rely on well known image metrics to compare performance of individual warping methods. Since the goal is to evaluate efficacy in compensating human body motion and not capacity for learning the background scenery, we use foreground masks for computing the image metrics. To this goal, we compute human body masks using a pre-trained image segmentation model [11]. Then, we filter out the background pixels for the PSNR metric. For the structural metrics of SSIM and LPIPS [12], we set the background pixels to zero in both the predicted and ground-truth images.

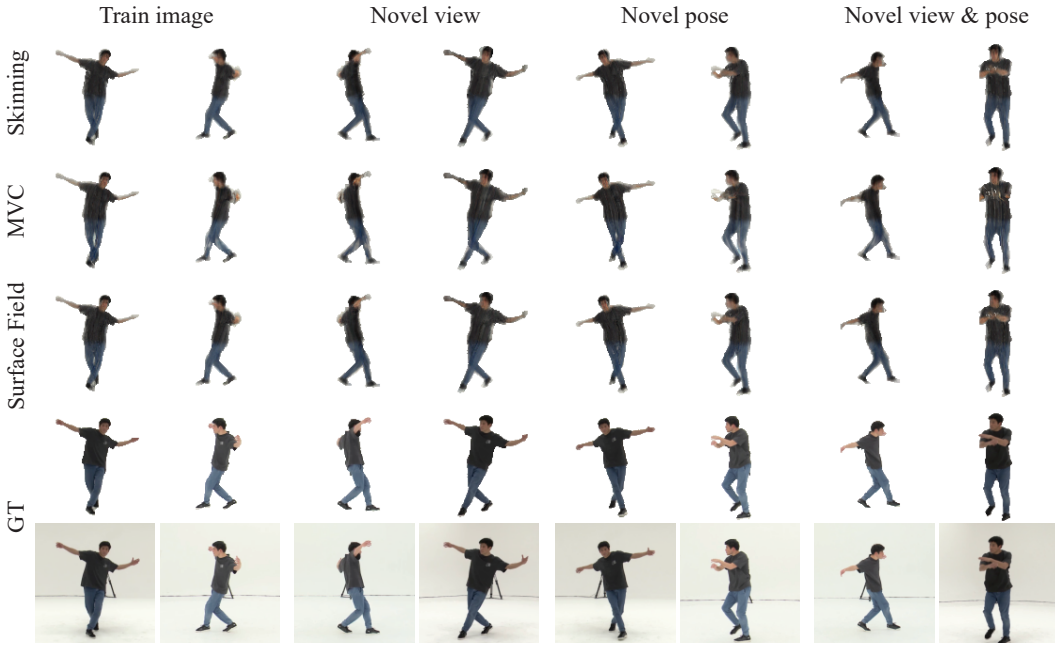

Figure 2: Qualitative comparison of results from our single-scene overfitting experiment. The figure presents the original training views and poses, interpolation of novel views, interpolation of novel poses and interpolation of novel poses under novel views. All images except the last row are presented with the same foreground masks as used for the metric evaluation.

## 2.2 Human body generation and animation

**SURREAL data pre-processing.** For training, we use the official SURREAL training split, and extract the first frame from each video. We square-crop each frame from head-to-toe using the ground truth segmentation mask, and resize the image to $128 \times 128$. The backgrounds are set to be black. We use the provided camera poses and intrinsics from the SURREAL dataset for training. Images for which the SMPL mesh scale is not consistent with the image size are filtered, leaving 35,332 total images for training.

The SURREAL dataset provides ground truth SMPL parameters for each frame, which we use. We compute the mean SMPL parameters across the frames in the dataset in order to assign the canonical pose, such that it is close to each of the target SMPL poses.

**AIST++ data pre-processing.** For training, we extract 30 frames uniformly sampled from each video in the AIST++ dataset. We then filter out frames whose effective camera distance to a normalized SMPL model is above a threshold or the projected human bounding box is partially outside of the image as a form of heuristic detecting poorly estimated SMPL parameters. We square-crop these frames at $600 \times 600$ resolution centered at the pelvis joint of the SMPL mesh, and resize each image to $256 \times 256$. Since ground truth masks are not provided for this dataset, we run an off-the-shelf segmentation model [11] to remove backgrounds. This stabilizes the GAN training, as the GAN no longer has to attempt to model a 3D-consistent background.

The AIST++ dataset also provides ground truth camera and body pose parameters. We move the translation of the SMPL mesh into the camera extrinsics, and simulate the scaling of the SMPL mesh by either moving the camera further back or closer. We additionally rescale all meshes and camera parameters such that the mean distance from camera to mesh is 1.7. Similarly to SURREAL, we compute the mean SMPL parameters across the selected training frames to assign the canonical pose.

**Evaluation.** We use the FID metric [13] for evaluation of the quality of generated images. This metric compares the distribution of intermediate features extracted from an inception network run on both generated and ground truth images. FID (10k) refers to the evaluation consistent with [14], which compares the distribution of 10,000 generated images with 10,000 randomly sampled ground truth images. FID (50k) compares the distribution of 50,000 generated images with the entire training dataset, giving a better estimate of the true distance. For EG3D, FID is run on the generated outputs with no warping. For the EG3D + warping baseline, the generated results from EG3D (not in the canonical pose) are warped by estimating the pose of the generated body and using this as the canonical pose. The FID is then applied to images generated in this fashion. For GNARF, we simply apply the FID to generated images with our deformation method.

In order to measure deformation accuracy for our method and the EG3D + warping method, we use the PCKh@0.5 metric. The use of this metric was inspired by the concurrent work [14]. After correspondence with the authors of this work, we have standardized our evaluation of this metric in order to compare reported values in the main paper table. Details of the evaluation method have a large effect on the magnitude of the PCKh@0.5 numbers reported, but describe the correct trend within a consistent evaluation standardization. For both the AIST++ and SURREAL dataset, we compute the GT keypoints by running a off-the-shelf body keypoint estimator [15] trained on MPII [16] (publicly available on the MMPose Project [17]) on each GT image. We then generate an image and deform the generated result with the body-pose parameters corresponding to this GT image, and render the generated result from the same camera position. The keypoint estimator is then run on this generated image, and the keypoints detected are compared in 2D. We discard keypoints that the keypoint estimator is not confident on in order to factor out estimator error, and only compare confident keypoints. We determine the head size (interocular distance) using the detected keypoints from the GT image.

## 2.3 Human face generation and editing

Here, we outline the implementation differences in the human body application.

**Deformation.** We use the FLAME head model to drive the deformation. The original template FLAME [18] template model has 5023 vertices and 9976 faces. This mesh contains 3 unconnected

parts, modeling the base face and the two eyeballs respectively. Since there is no suitable method to accurately extract the FLAME parameters related to eye movements from training images, we remove the eyeball parts. In original template, the neck and mouth are modeled with holes. We find having holes slightly degrades the deformation quality for points around the hole area, likely because worse point-to-triangle correspondence. Furthermore, we find the small triangles can lead to numeric instabilities when e.g. computing the barycentric coordinates for deformation, therefore we decimate the mesh as described in the main paper which makes the triangle sizes more uniform and also speeds up the SF deformation. The resulting mesh template, shown in Fig. 3 contains $1,252$ vertices and $2,500$ triangles.

The default FLAME head model is in a different scale and origin as the head generated by the pretrained EG3D model. To ensure meaningful warping in the transfer learning, we rescale by $2.6$ and fix the root joint at $[-0.0013, -0.1344, -0.0390]^\mathsf{T}$, which we determined by visually aligning the heads from FLAME and a pretrained EG3D.

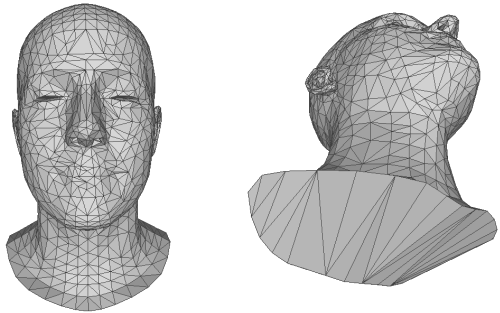

Figure 3: Processed FLAME model for training GNARF on FFHQ.

**Generator pose conditioning.** Unlike for the body, we use camera pose conditioning for the generator as proposed by EG3D. EG3D found that by using swapping regularization, camera pose conditioning will not negatively impact 3D consistency but rather improve generation quality. We found this true in our experiments for faces.

However, same as the body experiment, we do not provide the generator with any information related to the deformation (i.e. expression, shape and jaw rotation). As explained before, this is crucial for generating consistent canonical faces.

**Volume Rendering.** Unlike for bodies, the final resolution of the output is $512 \times 512$, consistent with the original EG3D.

**Dataset preprocessing.** Our data preparation is based on the original EG3D [1]. For each training image, we fix the camera intrinsics and estimate the camera extrinsics assuming that the head is inside a unit-length bounding box, front-facing and at a fixed position. Additionally, we use DECA [19] to estimate the expression, pose (only the jaw rotation, since we assume that the head is front-facing) and the shape parameters of the FLAME models. Note that, originally, DECA also outputs the camera parameters. However it uses an orthographic camera model, which is not directly transferable to the camera used in the pretrained EG3D. We therefore use the camera pose estimation from EG3D's data preprocessing procedure.



\section*{Checklist}

\begin{enumerate}

\item For all authors...
\begin{enumerate}
  \item Do the main claims made in the abstract and introduction accurately reflect the paper's contributions and scope?
    \answerYes{}
  \item Did you describe the limitations of your work?
    \answerYes{See Section~\ref{sec:discussion}.}
  \item Did you discuss any potential negative societal impacts of your work?
    \answerYes{See Section~\ref{sec:discussion}.}
  \item Have you read the ethics review guidelines and ensured that your paper conforms to them?
    \answerYes{}
\end{enumerate}

\item If you are including theoretical results...
\begin{enumerate}
  \item Did you state the full set of assumptions of all theoretical results?
    \answerNA{}
        \item Did you include complete proofs of all theoretical results?
    \answerNA{}
\end{enumerate}

\item If you ran experiments...
\begin{enumerate}
  \item Did you include the code, data, and instructions needed to reproduce the main experimental results (either in the supplemental material or as a URL)?
    \answerNo{We plan to release the code completely, but have not yet with submission. Details on the architecture, training, and data are available in the supplemental material.}
  \item Did you specify all the training details (e.g., data splits, hyperparameters, how they were chosen)?
    \answerYes{See supplemental document and sections~\ref{sec:experiment_body} and \ref{sec:experiment_faces}.}
        \item Did you report error bars (e.g., with respect to the random seed after running experiments multiple times)?
    \answerNo{\moniker{} models require a significant amount of computational resources (see supplemental document), and thus averaging training over many seeds would be computationally infeasible. However, we evaluate our trained models with a large number of generated images, demonstrating the robustness of our method.}
        \item Did you include the total amount of compute and the type of resources used (e.g., type of GPUs, internal cluster, or cloud provider)?
    \answerYes{See supplemental document.}
\end{enumerate}

\item If you are using existing assets (e.g., code, data, models) or curating/releasing new assets...
\begin{enumerate}
  \item If your work uses existing assets, did you cite the creators?
    \answerYes{}
  \item Did you mention the license of the assets?
    \answerYes{See Section~\ref{sec:experiments}.}
  \item Did you include any new assets either in the supplemental material or as a URL?
    \answerNo{}
  \item Did you discuss whether and how consent was obtained from people whose data you're using/curating?
    \answerNA{}
  \item Did you discuss whether the data you are using/curating contains personally identifiable information or offensive content?
    \answerNA{}
\end{enumerate}

\item If you used crowdsourcing or conducted research with human subjects...
\begin{enumerate}
  \item Did you include the full text of instructions given to participants and screenshots, if applicable?
    \answerNA{}
  \item Did you describe any potential participant risks, with links to Institutional Review Board (IRB) approvals, if applicable?
    \answerNA{}
  \item Did you include the estimated hourly wage paid to participants and the total amount spent on participant compensation?
    \answerNA{}
\end{enumerate}

\end{enumerate}

\end{document}